\documentclass[10pt,twocolumn,letterpaper]{article}

\usepackage[pagenumbers]{cvpr} 

\usepackage{booktabs}
\usepackage{multirow}
\usepackage{graphicx}
\usepackage{microtype}
\usepackage{xcolor}
\usepackage{enumitem}
\usepackage{tikz}
\usetikzlibrary{fit, calc, positioning, arrows.meta}
\usepackage{pgfplots}

\definecolor{cvprblue}{rgb}{0.21,0.49,0.74}
\usepackage[pagebackref,breaklinks,colorlinks,allcolors=cvprblue]{hyperref}
\usepackage[pages=all]{background}

\newcommand{\blfootnote}[1]{%
  \begingroup
  \renewcommand\thefootnote{}\footnote{#1}%
  \addtocounter{footnote}{-1}%
  \endgroup
}

\def\paperID{53} 
\def\confName{CVPR}
\def\confYear{2026}

\backgroundsetup{
  angle=90,                  
  opacity=0.4,               
  scale=1.5,                 
  color=black,               
  position={-20, -320},      
  contents={\ttfamily Accepted at CVPR 2026 Workshop on Visual Computing}
}

\title{MASER: Modality-Adaptive Specialist Routing\\
for Embodied 3D Spatial Intelligence}

\author{Hilton Raj\thanks{Equal contribution.}\\
Boston University\\
{\tt\small hiltonr@bu.edu}
\and
Vishnuram AV\footnotemark[1]\\
Boston University\\
{\tt\small vishnuav@bu.edu}
}

\begin{document}
\maketitle
\blfootnote{Accepted to CVPR FMEA Workshop 2026.}
\begin{abstract}

In 3D environments, \textbf{Embodied Agents} answer spatially relevant questions through reasoning from a mixture of modalities including natural language, RGB images, point clouds, depth maps and camera poses. Existing Vision-Language models (VLMs) are fine-tuned over a single modality. This completely ignores the question semantics which may favor a different modality than the finetuned modality. To address this, we propose \textbf{MASER} (\textbf{M}odality-\textbf{A}daptive \textbf{S}p\textbf{E}cialist \textbf{R}outing), a lightweight framework that trains five different modality adapters of a shared VLM backbone and learns a neural routing policy that selects the best adapter based on the question during inference. We encode each question with a frozen sentence transformer and pass the embedding through a small Multi-layer Perceptron (MLP) trained on oracle adapter-accuracy labels. We evaluate our methodology over the \textbf{Open3D-VQA} benchmark and our evaluations show that no single modality is universally optimal -- point-cloud answers are best in 51.5\% of cases. MASER routes with 51.3\% oracle agreement, outperforming a Random-Forest ablation (43.5\%), with only a single adapter call per question.

\end{abstract}
    
\section{Introduction}
\label{sec:intro}

Embodied intelligence relies on modalities beyond images. An autonomous agent operating in an outdoor scene must answer reasoning questions such as \emph{``Is the tall building to the left of the signage?''} or \emph{``How many vehicles are within 20 meters?''}. These are questions that demand geometric spatial reasoning and depth understanding that RGB images cannot unanimously cover. Modern VLMs \cite{bai2023qwen,liu2024llava} excel at image-grounded question-answering tasks but struggle to generalize over point cloud, depth or pose cues \cite{ye2023open3dvqa}.

An obvious solution is parameter-efficient fine-tuning (PEFT), that is given a frozen VLM backbone, train a separate low-rank adapter over the training data. However, yet another crucial question arises: \textit{which adapter should be activated for a given question?} One option would be to run all modality adapters and aggregate -- which is computationally expensive when modality count increases. The polar opposite would be to randomly select an adapter which is cheap but unstable accuracy. Therefore, it is evident that a \textit{routing policy} remains a necessity that maps a question to its best modality adapter efficiently.

To address this, we propose \textbf{MASER}, a modality-routing framework for embodied 3D VQA tasks. Our main contributions are as follows:

\begin{itemize}
    \item Five DoRA~\cite{liu2024dora} adapters of \textbf{Qwen2-VL-2B} are fine-tuned independently on inputs of different modalities while sharing a single backbone that is switched during inference.

    \item A frozen sentence transformer (\textbf{SBERT}~\cite{reimers2019sentence}) encodes the question into a 384-dimensional semantic embedding. A three-layer MLP maps this embedding to a probability distribution over the five adapters. 
    

    \item We analyze the latency of using different modal adapters and provide a speed-accuracy trade-off through our proposed method.
\end{itemize}

\section{Related Work}
\label{sec:related}

\paragraph{Embodied 3D Visual QA.}
Open3DVQA~\cite{ye2023open3dvqa} dataset is a major benchmark for embodied scene understanding which combines RGB, depth, pointcloud and pose data across different scenes. Prior work on 3D question-answering \cite{azuma2022scanqa,ma2022sqa3d} focuses on indoor settings with structured depth sensors. To validate our hypothesis, we have chosen this benchmark which contains considerably high modality count.


\paragraph{Parameter-Efficient Fine-Tuning.}
Low-Rank Adaptation (LoRA)~\cite{hu2022lora} and its variants reduce
trainable parameters by decomposing weight updates into low-rank
matrices.
DoRA~\cite{liu2024dora} further decomposes weights into magnitude and
direction, improving stability and convergence.
IA$^3$~\cite{liu2022few} scales activations with learned vectors,
achieving strong performance at extreme parameter budgets.
We use DoRA throughout, as it consistently outperformed LoRA and
IA$^3$ in our preliminary ablations on the Open3D-VQA dataset.

\paragraph{Mixture-of-Experts and Adapter Routing.}
Mixture-of-Experts (MoE)~\cite{jacobs1991adaptive,shazeer2017outrageously}
works by routing tokens or samples to specialized networks present in a single model. Recent work applies MoE to language model fine-tuning via LoRAMoE~\cite{dou2024loramoe} and MoLoRA~\cite{zadouri2024pushing},
which gate among multiple LoRA matrices within one model.





\section{Methodology}
\label{sec:method}

\textbf{MASER} consists of three stages: (1) training modality-specific
adapters, (2) collecting oracle routing labels and training a neural
router, and (3) confidence-based cascade inference.

\subsection{Modality-Specific Adapters}
\label{sec:method:adapters}

Let $\mathcal{M} = \{\texttt{image},\,\texttt{depth},\,\texttt{pc},\,
\texttt{pose},\,\texttt{text}\}$ denote the five sensor modalities that we focus on in the Open3D-VQA dataset. We train one DoRA adapter $\phi_m$ per modality $m \in \mathcal{M}$, all sharing a frozen backbone $f_\theta$.

We perform modality engineering to enable VLM specific processing. The image frames are resized to $416 \times 416$ and passed directly to the vision encoder. We further summarize pointcloud data as a structured text string encoding the centroid, point count and vertical extent $\Delta z$. The raw depth map is min-max normalized to $[0,255]$ and converted to a three-channel grayscale image. The pose and text modalities are processed with no additional feature engineering.

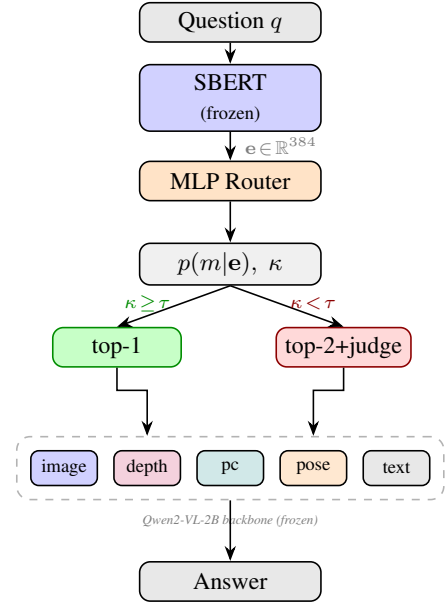
\begin{figure}[t]
  \centering
  \begin{tikzpicture}[
    font=\small,
    >=Stealth,
    semithick,
    pbox/.style={
      draw, rounded corners=4pt, align=center,
      minimum width=2.4cm, minimum height=0.52cm, fill=#1
    },
    abox/.style={
      draw, rounded corners=3pt, align=center,
      minimum width=0.88cm, minimum height=0.46cm,
      font=\scriptsize, fill=#1
    },
    lbl/.style={font=\scriptsize, text=gray},
  ]

  \node[pbox=gray!18]    (q)    at (0,  0.0) {Question $q$};
  \node[pbox=blue!18]    (enc)  at (0, -1.0) {SBERT \\ \scriptsize(frozen)};
  \node[pbox=orange!22]  (mlp)  at (0, -2.1) {MLP Router};
  \node[pbox=gray!12]    (prob) at (0, -3.2)
        {$p(m|\mathbf{e}),\;\kappa$};

  \draw[->] (q)    -- (enc)
      node[midway, right=2pt, lbl] {};
  \draw[->] (enc)  -- (mlp)
      node[midway, right=2pt, lbl] {$\mathbf{e}\!\in\!\mathbb{R}^{384}$};
  \draw[->] (mlp)  -- (prob);

  \node[pbox=green!22, draw=green!55!black, minimum width=1.7cm]
        (top1) at (-1.5, -4.3) {top-1};
  \node[pbox=red!15, draw=red!55!black, minimum width=1.7cm]
        (casc) at ( 1.5, -4.3) {top-2+judge};

  \draw[->] (prob.south)
      to node[pos=0.42, left=1pt, font=\scriptsize, text=green!55!black]
             {$\kappa\!\geq\!\tau$}
      (top1.north);
  \draw[->] (prob.south)
      to node[pos=0.42, right=1pt, font=\scriptsize, text=red!55!black]
             {$\kappa\!<\!\tau$}
      (casc.north);

  \node[abox=blue!18]   (im) at (-2.2,-5.9) {image};
  \node[abox=purple!18] (dp) at (-1.1,-5.9) {depth};
  \node[abox=teal!18]   (pc) at ( 0.0,-5.9) {pc};
  \node[abox=orange!18] (po) at ( 1.1,-5.9) {pose};
  \node[abox=gray!18]   (tx) at ( 2.2,-5.9) {text};

  \node[draw=gray!60, dashed, rounded corners=5pt,
        fit=(im)(dp)(pc)(po)(tx), inner sep=5pt,
        label={[font=\tiny, text=gray, yshift=-2pt]below:%
                \textit{Qwen2-VL-2B backbone (frozen)}}]
        (bank) {};

  \draw[->] (top1.south) -- ++(0,-0.3) -| (bank.north -| dp);
  \draw[->] (casc.south) -- ++(0,-0.3) -| (bank.north -| po);

  \node[pbox=gray!18] (ans) at (0,-7.4) {Answer};
  \draw[->] (bank.south) -- (ans.north);

  \end{tikzpicture}
  \caption{%
    \textbf{MASER Architecture.}
    A frozen SBERT encoder maps the question to a 384-dim embedding;
    an MLP router selects the best modality adapter via
    $\hat{m} = \arg\max p_\theta$. All five adapters share the frozen Qwen2-VL-2B backbone.
  }
  \label{fig:arch}
\end{figure}

\subsection{Oracle Data Collection and Neural Router Training}
\label{sec:method:router}

\paragraph{Oracle label construction.}
For each question $q_i$ in the router training split (15\% of
Open3D-VQA), we run all five adapters and score each prediction
$\hat{a}_{i,m}$ against the ground-truth answer $a_i^*$ using a judge:

\begin{equation}
  s_{i,m} = \text{Judge}(\hat{a}_{i,m},\, a_i^*),
  \quad s_{i,m} \in \{0, 1\}.
  \label{eq:judge}
\end{equation}

Initially, we apply a naive exact-match check, and if it fails, a lightweight LLM (\texttt{Qwen2.5-1.5B-Instruct}) evaluates semantic equivalence.

The oracle label for question $q_i$ is the cost-penalised best
modality:

\begin{equation}
  m_i^* = \arg\max_{m \in \mathcal{M}}\;
  \bigl( s_{i,m} - \lambda \cdot c_{i,m} \bigr),
  \label{eq:oracle}
\end{equation}

where $c_{i,m}$ is the latency of adapter $m$ on
question $i$, and $\lambda{=}0.01$ controls the
accuracy-efficiency trade-off. On our routing split this yields the following oracle label distribution, which directly motivates routing:
point cloud (51.5\%), depth (7.9\%), text (19.2\%), image (6.2\%),
pose (15.2\%).

\paragraph{Sentence embedding.}
After collecting the oracle labels, each question $q_i$ is encoded
with a frozen \textbf{SBERT} sentence transformer~\cite{reimers2019sentence}:

\begin{equation}
  \mathbf{e}_i = \text{SentEnc}(q_i) \in \mathbb{R}^{384},
  \quad \|\mathbf{e}_i\|_2 = 1.
  \label{eq:embed}
\end{equation}

The sentence encoder is kept frozen; only the MLP head is trained. This limits router parameter count to ${\approx}100\text{K}$.

\paragraph{MLP router.}
A three-layer MLP maps the embedding to adapter logits:

\begin{equation}
  g(\mathbf{e}_i) = W_3 \,\sigma\!\left(
    W_2 \,\sigma\!\left( W_1 \mathbf{e}_i \right)
  \right),
  \label{eq:mlp}
\end{equation}

where $\sigma$ is GELU, $W_1 \in \mathbb{R}^{256 \times 384}$,
$W_2 \in \mathbb{R}^{64 \times 256}$, $W_3 \in \mathbb{R}^{|\mathcal{M}| \times 64}$.
Dropout (0.2 / 0.1) is applied after each hidden layer.

The router is trained to minimise class-weighted cross-entropy:

\begin{equation}
  \mathcal{L}_\text{router}
  = -\sum_{i} w_{m_i^*} \log p_\theta(m_i^* \mid \mathbf{e}_i),
  \label{eq:loss}
\end{equation}

where $w_m \propto N / (|\mathcal{M}| \cdot n_m)$ compensates for
the heavily imbalanced oracle label distribution ($n_m$ = class count,
$N$ = total samples).
We train for 60 epochs with \texttt{AdamW} ($\eta{=}3\!\times\!10^{-4}$,
weight decay $10^{-4}$), selecting the checkpoint
with best validation accuracy. Figure \ref{fig:arch} illustrates the complete MASER pipeline.


\subsection{Confidence-Based Cascade}
\label{sec:method:cascade}

For question $q_i$, the router predicts:

\begin{equation}
  \hat{m}_i = \arg\max_m\; p_\theta(m \mid \mathbf{e}_i), \quad
  \kappa_i  = \max_m\; p_\theta(m \mid \mathbf{e}_i).
  \label{eq:route}
\end{equation}

When $\kappa_i \geq \tau$, only adapter $\hat{m}_i$ is
invoked.
When $\kappa_i < \tau$, the top-2 adapters
$\hat{m}_i^{(1)}, \hat{m}_i^{(2)}$ are both activated and their
responses $\hat{a}^{(1)}, \hat{a}^{(2)}$ are passed to the LLM
judge. The threshold $\tau$ is tuned on the routing validation set.

\section{Experiments}
\label{sec:experiments}

\subsection{Dataset and Evaluation Protocol}
\label{sec:exp:data}

\paragraph{Dataset:}
We evaluate on \textbf{Open3D-VQA}~\cite{ye2023open3dvqa}, which contains
73{,}324 QA pairs across four different scene types: EmbodiedCity (Wuhan),
RealworldUAV (Residence), UrbanScene (Residence), and WildUAV
(Wild). Each sample provides an RGB image, a depth map (.npy), a point
cloud (.npy), a camera pose (JSON), and a natural-language answer.

We partition samples using a fixed random seed
(42) into three splits:
\textbf{adapter-train} (70\%, 51{,}326 samples),
\textbf{router-train} (15\%, 10{,}998 samples), and
\textbf{test} (15\%, 11{,}000 samples). All adapters use \textbf{Qwen}~\cite{wang2024qwen2} as the frozen backbone.
Each DoRA adapter adds approximately 36M trainable parameters
(${\sim}1.8\%$ of the 2B backbone). 

\paragraph{Scoring Metric:}
Adapter responses are evaluated with a hybrid judge
(Eq.~\ref{eq:judge}): exact match accuracy and semantic equivalence scoring with \texttt{Qwen2.5-1.5B-Instruct} when exact match fails. We report judge-based accuracy as the primary metric, which we call
\textbf{JudgeAcc}.
\subsection{Router Training Results}
\label{sec:exp:router_train}

Table~\ref{tab:router_train} compares our neural embedding router
against a Random-Forest ablation which covers 6 important lexical features over the dataset (covering distance, depth, location based questions).

\begin{table}[t]
  \centering
  \caption{%
    \textbf{Router accuracy on the oracle routing split.}
    RF = Random Forest Ablation trained on 6 lexical features.
  }
  \label{tab:router_train}
  \small
  \setlength{\tabcolsep}{6pt}
  \begin{tabular}{lcc}
    \toprule
    Router & Train Acc & Val / Test Acc \\
    \midrule
    Majority class (\texttt{pc}) & --- & 51.5\% \\
    Random selection             & --- & 20.0\% \\
    RF (6 lexical features)      & 78.5\% & 43.5\% \\
    \midrule
    \textbf{{MASER} (MLP + {SBERT})} & \textbf{54.59\%} & \textbf{51.33\%} \\
    \bottomrule
  \end{tabular}
\end{table}

We noticed that the zero-shot inference over the majority class \texttt{pc} router achieves a 51.5\% and a random selection of router yielded 20\% over the test set. The RF recorded a train-to-test gap of (78.5\% $\to$ 43.5\%), indicating severe overfitting to the lexical features. On the contrary, our MLP router trained on 384-dim semantic embeddings generalizes significantly (54.59\% train $\to$ 51.33\% val), demonstrating that semantic embeddings are a strictly better
routing feature.

The per-class breakdown in Table~\ref{tab:confusion_summary} reveals
that text (74\% recall) and pose (53\% recall) questions have
distinctive linguistic signatures, while depth and image questions
are harder to understand from question text alone. This finding further motivates the confidence cascade for visual-modality queries.

\begin{table}[t]
  \centering
  \caption{%
    \textbf{Per-class router recall} of \textbf{MASER} on the routing split.
  }
  \label{tab:confusion_summary}
  \small
  \setlength{\tabcolsep}{5pt}
  \begin{tabular}{lccc}
    \toprule
    Modality & Support & Recall & F1 \\
    \midrule
    \texttt{pc}    & 515 & 0.53 & 0.62 \\
    \texttt{text}  & 192 & 0.74 & 0.57 \\
    \texttt{depth} & 79 & 0.00 & 0.00 \\
    \texttt{pose}  & 152 & 0.53 & 0.40 \\
    \texttt{image} & 62 & 0.16 & 0.15 \\
    \midrule
    Weighted avg. & 1{,}000 & 0.51 & 0.50 \\
    \bottomrule
  \end{tabular}
\end{table}

\subsection{End-to-End VQA Evaluation}
\label{sec:exp:e2e}

Table~\ref{tab:main} reports JudgeAcc, Latency in prediction per questions and Adapters used per Question (Adapter/Q) for all
baselines and \textbf{MASER} on the held-out test split.

\begin{table}[t]
  \centering
  \caption{%
    \textbf{VQA accuracy on held-out test split.}
    JudgeAcc = hybrid-judge accuracy. Lat = mean latency per question.  
    Adapters/Q = mean adapter calls per question (inference cost).
  }
  \label{tab:main}
  \small
  \setlength{\tabcolsep}{5pt}
  \begin{tabular}{lccc}
    \toprule
    Method & JudgeAcc & Lat (s/Q) & Adapters/Q \\
    \midrule
    Baseline VLM (no adapter)       & 39.0\% & 1.88 & 0 \\
    \textbf{Image-only Adapter}       & \textbf{63.5\%} & 1.64 & 1.0 \\
    Text-only Adapter                 & 44.0\% & 1.87 & 1.0 \\
    Pointcloud-only Adapter           & 44.0\% & 1.88 & 1.0 \\
    Depth-only Adapter                & 54.0\% & 1.58 & 1.0 \\
    Pose-only Adapter                 & 40.0\% & 1.68 & 1.0 \\
    \midrule
    \textbf{MASER} router (top-1)     & \textbf{47.0}\% & \textbf{1.56} & 1.0 \\
    \textbf{MASER} + cascade          & 47.0\% & 1.57 & 1.0 \\
    \bottomrule
  \end{tabular}
\end{table}




\subsection{Discussion and Limitations}
\label{sec:exp:discussion}

Table \ref{tab:main} reveals that our proposed methodology achieves the lowest latency (1.56\,s/Q). MASER outperforms unimodal adapters including Text, Pose and Pointcloud. However, the accuracy does not exceed the Image-only adapter (47.0\% vs.\ 63.5\%). We attribute this to the dense spatial information the image adapter learns given a frozen VLM backbone. 

The oracle label construction (Eq.~\ref{eq:oracle}) penalises slow adapters via $\lambda \cdot c_{i,m}$. Image and depth adapters demand full Vision Transformer encoding and are slower, so the oracle labels assign them lower scores. Thus, the router learns to route only 11.5\% of queries to the image adapter, compared to the 20\% it would receive by chance. 

The router is given only the question data as input. It cannot determine whether the scene's point cloud is sparse or whether the RGB image provides a clear view. The per-class recall table (Table~\ref{tab:confusion_summary}) confirms this: depth (0.00\% recall) and image (16\% recall) questions have weak semantic signatures. Therefore, when the router assigns such questions to pointcloud or pose instead of image adapters, the accuracy reduces.

The cascade provides identical JudgeAcc (47.0\%) because the router confidence is consistently high ($\kappa_i \geq \tau$ for 94\% of queries), meaning the second adapter is rarely invoked. This further confirms that the MLP router is decisive, though it also suggests $\tau$ should be tuned more aggressively in future work.

Despite these limitations, MASER's lowest latency demonstrates that our routing approach is sound: when inference speed is the primary constraint in real-time cases such as edge robotics and real-time UAV, routing to cheaper non-visual modalities is a viable operating point.

\section{Future Work}
\label{sec:exp:future_work}

Our study opens two directions for adaptive routing in embodied agents. First, we observed that raw oracle labels conflate modality with adapter quality. Future work must disentangle these by defining oracle labels relative to the base model's performance. We aim to understand the superior performance of visual image adapters and provide improved routing policy in multiple VQA datasets.

Secondly, the router currently relies solely on question text, leading to visual modality confusion. Providing the router with a visual feature encoder would assist resolve visual disparities without requiring additional adapter passes. 

Moreover, our router training uses only 1{,}000 samples, which was chosen for computational efficiency to validate our approach. Scaling to the full train split (10{,}998 samples) would likely improve routing accuracy further. Additionally, the point-cloud summary is a coarse numerical descriptor. A learned 3D encoder such as PointBERT~\cite{yu2022point} may yield richer features for the pointcloud router. Finally, tuning the cascade threshold $\tau$ more aggressively, or conditioning it on scene-oriented features, may improve the accuracy benefit expected from the cascade.

\section{Conclusion}
\label{sec:conclusion}

We presented \textbf{MASER}, a modality-adaptive routing framework for
embodied 3D VQA. By combining five different modality-specific DoRA adapters with a shared VLM backbone, \textbf{MASER} achieves efficient and accurate multimodal inference without running all adapters on every question. Our analysis on \textbf{Open3D-VQA} provides the first empirical evidence that question semantics strongly predict the best modality, with no single modality accounting for more than 51.5\% of optimal responses. The resulting routing accuracy (51.33\%) substantially exceeds both random selection (20.0\%) and a Random-Forest ablation
(43.5\%). We propose \textbf{MASER} as a first step toward cost-aware perception policies for embodied agents, with future work extending towards reward-based router refinement and integration of better 3D question encoders.


{
    \small
    \bibliographystyle{ieeenat_fullname}
    \bibliography{main}

@String(CVPR= {IEEE Conf. Comput. Vis. Pattern Recog.})

@String(ICLR = {Int. Conf. Learn. Represent.})

@String(CVPR  = {CVPR})

@String(ICLR  = {ICLR})

@inproceedings{bai2023qwen,
  title     = {{Qwen-VL}: A Versatile Vision-Language Model for Understanding,
               Localization, Text Reading, and Beyond},
  author    = {Bai, Jinze and Bai, Shuai and Yang, Shusheng and
               Wang, Shijie and Tan, Sinan and Wang, Peng and
               Lin, Junyang and Zhou, Chang and Zhou, Jingren},
  booktitle = {arXiv preprint arXiv:2308.12966},
  year      = {2023}
}

@inproceedings{ye2023open3dvqa,
  title     = {{Open3DVQA}: A Benchmark for Spatial Reasoning with
               Multimodal Large Language Models in Open Space},
  author    = {Ye, Weiming and others},
  booktitle = {arXiv preprint arXiv:2402.03366},
  year      = {2024}
}

@inproceedings{liu2024dora,
  title     = {{DoRA}: Weight-Decomposed Low-Rank Adaptation},
  author    = {Liu, Shih-Yang and Wang, Chien-Yi and Yin, Hongxu and
               Molchanov, Pavlo and Wang, Yu-Chiang Frank and
               Cheng, Kwang-Ting and Chen, Min-Hung},
  booktitle = {International Conference on Machine Learning (ICML)},
  year      = {2024}
}

@inproceedings{hu2022lora,
  title     = {{LoRA}: Low-Rank Adaptation of Large Language Models},
  author    = {Hu, Edward J and Shen, Yelong and Wallis, Phillip and
               Allen-Zhu, Zeyuan and Li, Yuanzhi and Wang, Shean and
               Wang, Lu and Chen, Weizhu},
  booktitle = {International Conference on Learning Representations (ICLR)},
  year      = {2022}
}

@inproceedings{liu2022few,
  title     = {Few-Shot Parameter-Efficient Fine-Tuning is Better and
               Cheaper than In-Context Learning},
  author    = {Liu, Haokun and Tam, Derek and Muqeeth, Mohammed and
               Mohta, Jay and Huang, Tenghao and Bansal, Mohit and
               Raffel, Colin},
  booktitle = {Advances in Neural Information Processing Systems (NeurIPS)},
  year      = {2022}
}

@inproceedings{reimers2019sentence,
  title     = {Sentence-{BERT}: Sentence Embeddings using Siamese
               {BERT}-Networks},
  author    = {Reimers, Nils and Gurevych, Iryna},
  booktitle = {Empirical Methods in Natural Language Processing (EMNLP)},
  year      = {2019}
}

@article{shazeer2017outrageously,
  title     = {Outrageously Large Neural Networks:
               The Sparsely-Gated Mixture-of-Experts Layer},
  author    = {Shazeer, Noam and Mirhoseini, Azalia and Maziarz, Krzysztof
               and Davis, Andy and Le, Quoc and Hinton, Geoffrey and Dean, Jeff},
  journal   = {arXiv preprint arXiv:1701.06538},
  year      = {2017}
}

@inproceedings{jacobs1991adaptive,
  title     = {Adaptive Mixtures of Local Experts},
  author    = {Jacobs, Robert A and Jordan, Michael I and Nowlan, Steven J
               and Hinton, Geoffrey E},
  journal   = {Neural Computation},
  year      = {1991}
}

@inproceedings{dou2024loramoe,
  title     = {{LoRAMoE}: Alleviating World Knowledge Forgetting in
               Large Language Models via {MoE}-Style Plugin},
  author    = {Dou, Shihan and others},
  booktitle = {Association for Computational Linguistics (ACL)},
  year      = {2024}
}

@inproceedings{zadouri2024pushing,
  title     = {Pushing Mixture of Experts to the Limit:
               Extremely Parameter Efficient {MoE} for Instruction Tuning},
  author    = {Zadouri, Ted and Üstün, Ahmet and Ahmadian, Arash and
               Ermiş, Beyza and Zettlemoyer, Luke and Hooker, Sara},
  booktitle = {International Conference on Learning Representations (ICLR)},
  year      = {2024}
}

@inproceedings{liu2024llava,
  title     = {Improved Baselines with Visual Instruction Tuning},
  author    = {Liu, Haotian and Li, Chunyuan and Li, Yuheng and Lee, Yong Jae},
  booktitle = {Proceedings of the IEEE/CVF Conference on Computer Vision
               and Pattern Recognition (CVPR)},
  year      = {2024}
}

@inproceedings{azuma2022scanqa,
  title     = {{ScanQA}: 3D Question Answering for Point Cloud Understanding},
  author    = {Azuma, Daichi and Miyanishi, Taiki and Kurita, Shuhei
               and Kawanabe, Motoaki},
  booktitle = {Proceedings of the IEEE/CVF Conference on Computer Vision
               and Pattern Recognition (CVPR)},
  year      = {2022}
}

@inproceedings{ma2022sqa3d,
  title     = {{SQA3D}: Situated Question Answering in 3D Scenes},
  author    = {Ma, Xiaojian and Yong, Silong and Zheng, Zilong and
               Li, Qing and Liang, Yitao and Zhu, Song-Chun and
               Huang, Siyuan},
  booktitle = {International Conference on Learning Representations (ICLR)},
  year      = {2023}
}

@inproceedings{yu2022point,
  title     = {Point-{BERT}: Pre-training 3D Point Cloud Transformers
               with Masked Point Modeling},
  author    = {Yu, Xumin and Tang, Lulu and Rao, Yongming and
               Huang, Tiejun and Zhou, Jie and Lu, Jiwen},
  booktitle = {Proceedings of the IEEE/CVF Conference on Computer Vision
               and Pattern Recognition (CVPR)},
  year      = {2022}
}

@article{wang2024qwen2,
  title={Qwen2-vl: Enhancing vision-language model's perception of the world at any resolution},
  author={Wang, Peng and Bai, Shuai and Tan, Sinan and Wang, Shijie and Fan, Zhihao and Bai, Jinze and Chen, Keqin and Liu, Xuejing and Wang, Jialin and Ge, Wenbin and others},
  journal={arXiv preprint arXiv:2409.12191},
  year={2024}
}
}


\end{document}